\documentclass[conference]{IEEEtran}
\IEEEoverridecommandlockouts
\usepackage{cite}
\usepackage{amsmath,amssymb,amsfonts}
\usepackage{algorithmic}
\usepackage{graphicx}
\usepackage{textcomp}
\usepackage{xcolor}
\def\BibTeX{{\rm B\kern-.05em{\sc i\kern-.025em b}\kern-.08em
    T\kern-.1667em\lower.7ex\hbox{E}\kern-.125emX}}
\begin{document}

\title{EmoNeXt: an Adapted ConvNeXt for Facial Emotion Recognition}

\author{\IEEEauthorblockN{Yassine El Boudouri}
\IEEEauthorblockA{\textit{CESI LINEACT Laboratory} \\
\textit{UR 7527}\\
Dijon, 21800, France \\
yelboudouri@cesi.fr}
\and
\IEEEauthorblockN{Amine Bohi}
\IEEEauthorblockA{\textit{CESI LINEACT Laboratory} \\
\textit{UR 7527}\\
Dijon, 21800, France \\
abohi@cesi.fr}
}

\maketitle

\begin{abstract}
Facial expressions play a crucial role in human communication serving  as a powerful and impactful means to express a wide range of emotions. With advancements in artificial intelligence and computer vision, deep neural networks have emerged as effective tools for facial emotion recognition. In this paper, we propose EmoNeXt, a novel deep learning framework for facial expression recognition based on an adapted ConvNeXt architecture network. We integrate a Spatial Transformer Network (STN) to focus on feature-rich regions of the face and Squeeze-and-Excitation blocks to capture channel-wise dependencies. Moreover, we introduce a self-attention regularization term, encouraging the model to generate compact feature vectors. We demonstrate the superiority of our model over existing state-of-the-art deep learning models on the FER2013 dataset regarding emotion classification accuracy.

\end{abstract}


\section{Introduction}
Facial expressions are a powerful means of non-verbal communication in face-to-face interactions, allowing humans to convey a wide range of information. According to Albert Mehrabian, facial expressions are more effective than words in face-to-face communication \cite{mehrabian1968some}. He revealed that words contribute only 7\% to effective communication, while voice tone accounts for 38\% and body language for 55\%. Therefore, facial expressions play a vital role in human communication.

In this context, it is legitimate to attempt to model the process of human perception of facial expressions. In the last two decades, computer vision and artificial intelligence research have shown great interest in the automatic recognition of facial expressions in videos and static images. Facial Expression Recognition (FER) has gained immense importance in various fields such as security, healthcare, driver fatigue surveillance, and human-machine interaction applications \cite{cortellessa2021ai, sajjad2019raspberry, jeong2018driver}.

Over the past few years, numerous conventional FER approaches have emerged, employing classical descriptors to explicitly extract features from face data. These approaches can be categorized into two groups: geometric methods and appearance-based methods. While geometric features capture the shape, location and interconnections of facial components during expressions \cite{ghimire2017recognition, kotsia2006facial}, appearance-based features focus on the variations in facial appearance, such as wrinkles and furrows, and can be extracted from the whole face or specific regions \cite{sajjad2019raspberry, shan2009facial, chen2014facial}. To classify these features, encompassing both geometric and texture-based ones, various classifiers have been employed, including Support Vector Machines (SVM), K-Nearest Neighbor (KNN), as well as neural networks such as MultiLayer Perceptron (MLP).


Conventional FER approaches follow a two-step process: they initially analyze and define facial features, and subsequently utilize these features for inference. However, as these two steps are performed separately, sub-optimal performance is obtained, particularly when dealing with complex datasets containing numerous sources of variability. Consequently, It is more advantageous to perform these two steps together for better recognition performance.

Over the past two decades, deep neural networks have demonstrated exceptional effectiveness in automatic recognition tasks, making them a natural fit for automatic FER.
Deep learning is used to learn discriminative feature representations for automatic FER by designing a hierarchical architecture composed of multiple non-linear transformations based on different types of neural networks such as convolutional networks (CNNs) and recurrent networks (RNNs). These networks are coupled with a classification layer for the classification task. As a result, the learning parameters of the classifier are determined in conjunction with the learning of the feature representations. This automation of feature extraction and classification directly from raw data greatly reduces dependence on models based on face geometry and other preprocessing techniques.


In this paper, we introduce EmoNeXt, a novel deep learning framework for FER based on an adapted ConvNeXt network \cite{liu2022convnet}. We integrate a Spatial Transformer Network (STN) \cite{jaderberg2015spatial} to allow the network to learn and apply spatial transformations to input images and Squeeze-and-Excitation blocks \cite{hu2018squeeze} to enable adaptive recalibration of channel-wise feature. Furthermore, we combine a Self-Attention regularization term and the classical Cross-Entropy \cite{zhang2018generalized} Loss to encourage the model to generate compact features. The proposed architecture was able to produce significantly better results than the original ConvNeXt network and outperform other state-of-the-art deep learning models under the same experimental setup on the FER2013 dataset \cite{goodfellow2013challenges}.



\section{Related Work}

In this section, we will briefly review some recent research on facial emotion classification, with a specific focus on models that have been evaluated using the FER2013 dataset.

Given the remarkable achievements and the rise of deep learning in the realm of computer vision, particularly in image classification tasks, several studies have introduced a range of deep learning approaches to address automated FER on the FER2013 dataset with the sole objective of achieving the best possible classification accuracy.


In \cite{georgescu2019local}, Georgescu et al. presented a method where automatic features, learned by multiple CNN architectures, were combined with handcrafted features computed using the bag-of-visual-words (BOVW) model. Once the fusion of the two types of features is accomplished, a local learning framework is utilized to make predictions of the class label for each individual test image. 
In \cite{pecoraro2022local}, Pecoraro et al. introduced a novel channel self-attention module called Local multi-Head Channel (LHC), that can be seamlessly incorporated into any existing CNN architecture. This module offers a solution to the limitation of Global Attention mechanisms by effectively directing attention to crucial facial features that significantly influence facial expressions.
In another paper \cite{fard2022ad}, Fard et al. proposed an Adaptive Correlation (Ad-Corre) Loss consisting of three components: Feature Discriminator (FD), Mean Discriminator (MD) and Embedding Discriminator (ED). The proposed Ad-Corre Loss was combined with the classical Cross-Entropy Loss and used to train two backbone models: Xception \cite{chollet2017xception} and Resnet50 \cite{he2016deep}. The authors demonstrated that irrespective of the deep-learning model employed, the Ad-Corre Loss allows to increase the discriminative power of generated feature vectors, consequently leading to high accuracy in classification.
Another novel deep learning model known as Segmentation VGG-19, was proposed in a recent study by Vignesh et al. \cite{vignesh2023novel}. This model enhances FER by integrating U-Net \cite{ronneberger2015u} based segmentation blocks into the VGG-19 (Visual Geometry Group) architecture \cite{simonyan2014very}. By inserting these segmentation blocks between the layers of VGG-19, the model effectively emphasizes significant features from the feature map, leading to an enhanced feature extraction process.
In another recent paper \cite{mukhopadhyay2023deep}, Mukhopadhyay et al. presented a deep-learning-based FER method by exploiting textural features such as local binary patterns (LBP), local ternary patterns (LTP) and completed local binary patterns (CLBP). A CNN model is then trained over these textural image features to achieve improved accuracy in detecting facial emotions. 
In a recent publication by Shahzad et al. \cite{shahzad2023role}, a zoning-based FER (ZFER) was introduced. The objective of ZFER was to accurately identify additional facial landmarks such as eyes, eyebrows, nose, forehead, and mouth, enabling a more comprehensive understanding of deep facial emotions through zoning. After the initial steps of face detection and extraction of face landmarks, these zoning-based landmarks were employed to train a hybrid VGG-16 model. The resulting feature maps generated by the hybrid model were then utilized as input for a fully CNN to classify facial emotions.

Later in the results section, Table \ref{tab:accuracy_table} presents a compilation of references to other deep learning-based models utilized for the FER task. This table provides an overview of the classification scores achieved by these state-of-the-art FER models on the FER2013 dataset.

\section{Methods}
This section explores various techniques and components integrated into a comprehensive deep learning model designed specifically for facial emotion detection. The focus primarily revolves around three key aspects: the preprocessing of input images at the beginning of the network, the feature extraction and classification, and the loss function minimizations.

\subsection{Spatial Transformer Networks}
Spatial Transformer Networks (STN) \cite{jaderberg2015spatial} extend the concept of differentiable attention to encompass various spatial transformations. By integrating a differentiable geometric transformation module into the network architecture, STN allow neural networks to learn and apply spatial transformations to input data. This capability proves invaluable in FER, where the performance is heavily influenced by scale, rotation, and translation variations.

\begin{figure}[htbp]
  \centering
  \includegraphics[scale=0.3]
  {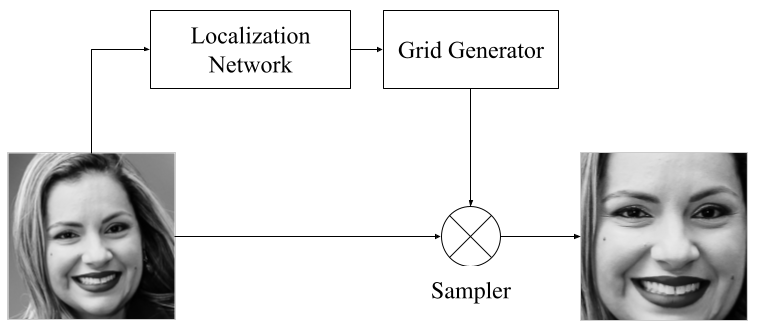}
  \caption{The architecture of a spatial transformer module.}
  \label{fig:stn_fig}
\end{figure}

The spatial transformer mechanism comprises three main components, illustrated in Figure \ref{fig:stn_fig}. The first component is the localization network, which employs convolutional layers to generate a vector representation of the input image. This vector is then utilized by the grid generator to create a sampling grid. The grid consists of points that determine where the input map should be sampled to generate the transformed output. Typically, fully connected layers are employed in the grid generator. Lastly, the feature map and the sampling grid are fed into the sampler, which samples the input image at the grid points to produce the final output image.

The key advantage of STN is their ability to learn the spatial transformations automatically as part of the neural network training process.

\subsection{ConvNeXt}
Introduced in 2022, ConvNeXt \cite{liu2022convnet} is a pure convolutional model that draws inspiration from Vision Transformers \cite{dosovitskiy2020image}. It was designed to compete with state-of-the-art Vision Transformers while retaining the simplicity and efficiency of CNNs. It incorporates various enhancements to the architecture of a standard ResNet \cite{he2016deep}, with many of these modifications evident in the ConvNeXt block.

\begin{figure}[htbp]
  \centering
  \includegraphics[scale=0.6]{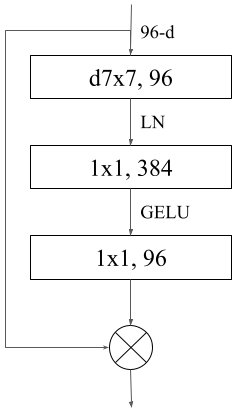}
  \caption{The ConvNeXt block.}
  \label{fig:convnext_block}
\end{figure}

The ConvNeXt block, illustrated in Figure \ref{fig:convnext_block}, uses larger kernel-sized and depthwise convolutions, increases the network width to 96 channels, and utilizes an inverted bottleneck design, to lower the overall network floating-point operations (FLOPs) while enhancing performance.

ConvNeXt also replaces ReLU \cite{nair2010rectified} with GELU \cite{hendrycks2016gaussian} as the activation function and substituting BatchNorm (BN) \cite{ioffe2017batch} with Layer Normalization (LN) \cite{ba2016layer} as the normalization technique, allowing the model to acheive a slightly better performance. GELU and LN are commonly used in advanced Transformers.

The authors of ConvNeXt have developed multiple versions, distinguished by variations in the number of channels (C) and blocks (B) within each stage. Here are the configurations:
\begin{align*}
\textbf{Tiny}   & \quad C = (96, 192, 384, 768) \quad & B = (3, 3, \phantom{0}9, 3) \\
\textbf{Small}  & \quad C = (96, 192, 384, 768) \quad & B = (3, 3, 27, 3) \\
\textbf{Base}   & \quad C = (128, 256, 512, 1024) \quad & B = (3, 3, 27, 3) \\
\textbf{Large}  & \quad C = (192, 384, 768, 1536) \quad & B = (3, 3, 27, 3) \\
\textbf{XLarge} & \quad C = (256, 512, 1024, 2048) \quad & B = (3, 3, 27, 3)
\end{align*}

\subsection{Squeeze-and-Excitation Blocks}
\begin{figure}[htbp]
  \centering
  \includegraphics[scale=0.5]{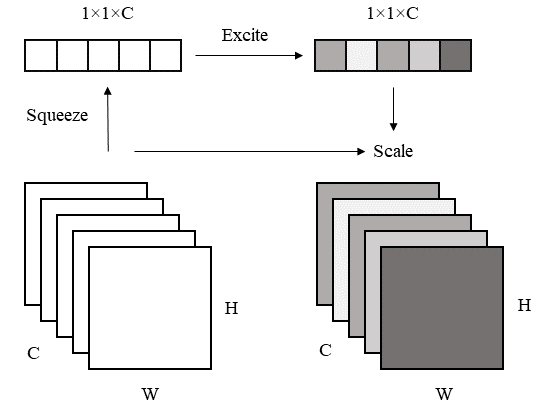}
  \caption{The architecture of the Squeeze-and-Excitation block.}
  \label{fig:se_block}
\end{figure}

Squeeze-and-Excitation (SE) \cite{hu2018squeeze} is a powerful technique used in deep learning models to enhance the representational power of CNN models. It introduces a mechanism that allows the network to adaptively recalibrate the channel-wise features, improving the model's discriminative capabilities. As illustrated in Figure \ref{fig:se_block}, the SE block consists of two fundamental operations: squeezing and exciting.
In the squeezing phase, global average pooling is applied to each channel of the feature map (W, H, C), reducing its spatial dimensions to a single value per channel (1, 1, C). The exciting phase follows, where the squeezed values are transformed using a small set of fully connected layers. These layers learn channel-wise weights, capturing the interdependencies among feature channels. The resulting attention weights are then multiplied element-wise with the original feature map, emphasizing informative channels and suppressing less relevant ones.

\subsection{EmoNeXt Final Architecture}
The EmoNeXt architecture begins with the inclusion of STN at the beginning of the network. STN enable the model to handle variations in scale, rotation, and translation by learning and applying spatial transformations to facial images.

After passing through the STN, the inputs are then passed through ConvNeXt's patchify module. This module downsamples the input image using a non-overlapping convolution with a kernel size of 4. The downsampling helps to reduce the dimensionality of the input and capture relevant features efficiently.

\begin{figure}[htbp]
  \centering
  \includegraphics[scale=0.7]{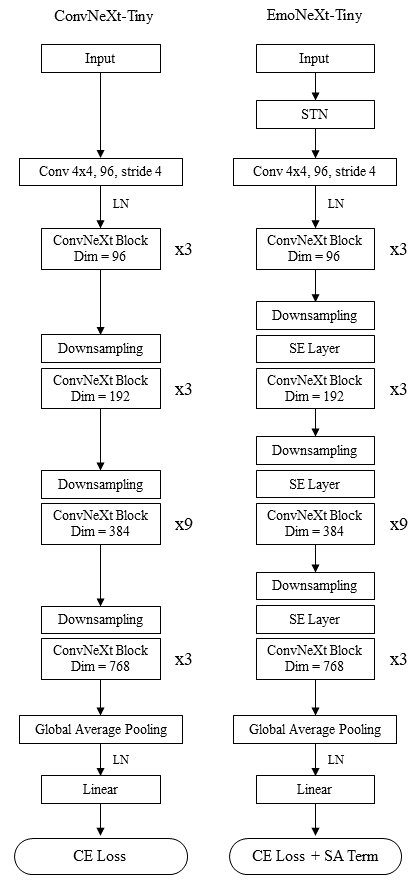}
  \caption{Architecture designes for ConvNeXt and EmoNeXt.}
  \label{fig:model_architecture}
\end{figure}

The downscaled inputs then go through the ConvNeXt stages. Each stage is followed by a SE block to recalibrate the feature map before going into the next stage. This recalibration enhances the model's ability to extract discriminative facial features for accurate emotion recognition. The overall architecture is illustrated in Figure \ref{fig:model_architecture}.

By leveraging these techniques, our EmoNeXt model achieves robust and accurate facial emotion detection, effectively handling variations in facial expressions and improving overall performance.

\subsection{Self-Attention Regularization Term}
Self-attention is a powerful mechanism used in various domains, including natural language processing, image analysis, and feature extraction \cite{vaswani2017attention}. It enables models to assign importance to different parts of the input sequence, allowing for effective capturing of dependencies and relationships.

A commonly used approach is the dot product self-attention, known for its simplicity and effectiveness. It calculates the relevance and importance of 
the feature vector by computing the dot product similarity between pairs of its elements. The resulting attention scores are then used to weight the feature vector, generating an attended representation that highlights the most significant information.

Mathematically, the dot product self-attention weights can be computed as follows:

\begin{equation}
\text{W}(\mathbf{Q}, \mathbf{K}) = \text{softmax}\left(\frac{\mathbf{Q} \cdot \mathbf{K}^\top}{\sqrt{d}}\right)
\end{equation}

Here, $\mathbf{Q}$ and $\mathbf{K}$ represent queries and keys of dimension $d$. We use a linear layer to compute the queries ($\mathbf{Q}$) and keys ($\mathbf{K}$) from a feature vector. The dot product between $\mathbf{Q}$ and $\mathbf{K}^\top$ computes pairwise similarities, while scaling the dot products by $\sqrt{d}$ prevents gradient issues. The softmax operation normalizes the attention scores.

To optimize the model's performance, an appropriate loss function is crucial. For multiclass classification tasks, the widely used Cross-Entropy Loss \cite{zhang2018generalized} serves as the standard choice. However, we introduce a novel component called the Self-Attention regularization term (SA).

The SA regularization term aims to minimize the attention weights to encourage balanced importance between generated features. It is defined as:

\begin{equation}
\mathcal{L}(W) = \frac{1}{N} \sum_{i=1}^{N} (W_i - \bar{W})^2
\end{equation}
where $W$ represents the self-attention weights, $\bar{W}$ represents the mean attention weights and $N$ is the length of the feature vector. By minimizing this term, we encourage the attention weights to be closer to the mean value, promoting balanced importance across the features.

The SA regularization term is added to the overall loss function of the model, in addition to the Cross-Entropy Loss. This helps to incorporate the regularization effect during training, encouraging the model to produce more balanced attention weights and improve generalization.

By combining the Cross-Entropy Loss and the SA regularization term, the final loss function becomes:

\begin{equation}
\mathcal{L}_\text{final} = \mathcal{L}_\text{CE} + \lambda \cdot \mathcal{L}_\text{SA}
\end{equation}
where $\mathcal{L}_\text{CE}$ is the Cross-Entropy Loss, $\mathcal{L}_\text{SA}$ is the SA regularization term, and $\lambda$ is a hyperparameter that controls the trade-off between the two terms.

\section{Experiments}
\subsection{Dataset}
\begin{figure}[htbp]
  \centering
  \includegraphics[scale=0.3]
  {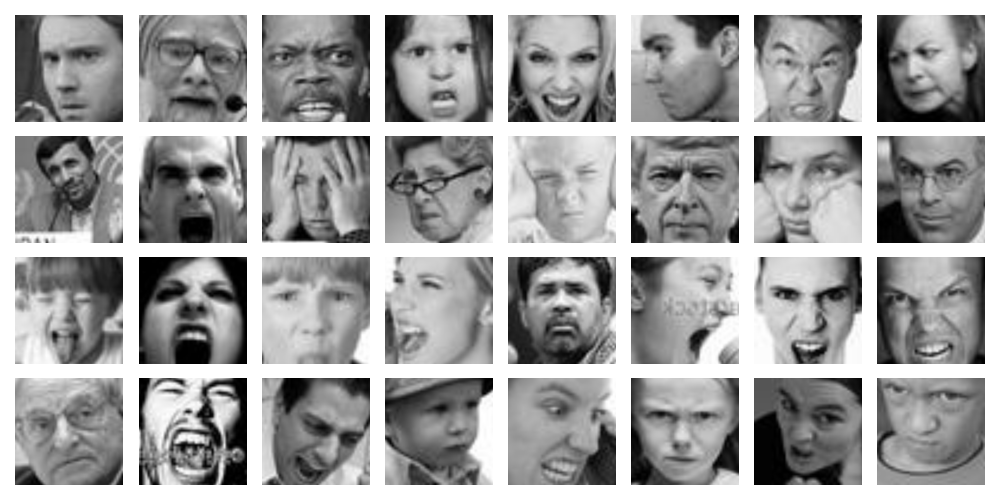}
  \caption{Sample images from the FER2013 dataset.}
  \label{fig:dataset_images}
\end{figure}

The experiments were conducted on the FER2013 dataset, initially introduced in ICML 2013 Challenges in Representation Learning \cite{goodfellow2013challenges}. The dataset consists of 35,887 grayscale images, each sized $48^2$ pixels (Figure \ref{fig:dataset_images}). It is divided into three subsets: 28,709 images for training, 3,589 images for validation, and 3,589 images for testing. The faces in the dataset are labeled with one of the seven classes as mentioned in Table \ref{tab:dataset_summary}. This dataset is widely used for FER tasks due to its significant number of samples. However, the distribution of classes in this dataset is highly unbalanced, which poses a challenge for any deep learning model. Figure \ref{fig:dataset_images} includes a selection of example data taken from FER2013. 

\begin{table}[h]
    \centering
    \caption{FER2013 dataset summary}
    \begin{tabular}{lcccc}
    \hline
    Class & Training & Validation & Testing & Class total \\
    \hline
    angry & 3995 & 467 & 491 & 4953 \\
    disgust & 436 & 56 & 55 & 547 \\
    fear & 4097 & 496 & 528 & 5121 \\
    happy & 7215 & 895 & 879 & 8989 \\
    sad & 4830 & 653 & 594 & 6077 \\
    surprise & 3171 & 415 & 416 & 4002 \\
    neutral & 4965 & 607 & 626 & 6198 \\
    Total & 28709 & 3589 & 3589 & 35887 \\
    \hline
    \end{tabular}
    \label{tab:dataset_summary}
\end{table}

\subsection{Training}
Recent studies have highlighted the effectiveness of modern training techniques in significantly improving the performance of deep learning models. In our model training, we employ various advanced strategies to improve the results. We utilize the AdamW optimizer \cite{loshchilov2017decoupled} with a learning rate of 1e-4, combined with a cosine decay schedule to enhance convergence. Additionally, we incorporate data augmentation techniques such as RandomCropping and RandomRotation to augment the training data, boosting the model's ability to generalize. To prevent overfitting, we implement regularization schemes like Stochastic Depth \cite{huang2016deep} and Label Smoothing \cite{szegedy2016rethinking}, which contribute to more robust and generalized models.

To further enhance the performance and address memory constraints, we employ the Exponential Moving Average (EMA) technique \cite{polyak1992acceleration}. EMA has proven effective in alleviating overfitting, particularly in larger models. Moreover, we adopt Mixed Precision Training \cite{micikevicius2017mixed}, a method that reduces memory consumption by almost 2x while accelerating the training process.

In addition, we enhance our model's capabilities by incorporating weights from pretrained ConvNeXt on the ImageNet-22k dataset \cite{russakovsky2015imagenet}. This dataset, known for its vast collection of diverse images, allows our model to leverage a wealth of learned knowledge. To ensure compatibility with the pretrained weights, we resize the images in our training pipeline to $224^2$, adhering to the established industry practice. This resizing technique enables us to effectively utilize the pretrained weights, resulting in improved performance and enhanced proficiency.

Finally, we trained both the EmoNeXt and ConvNeXt models, encompassing all five sizes (T, S, B, L, and XL), utilizing an Nvidia T4 GPU with 16GB of VRAM. The implementation was done using PyTorch version 2.0.0, and the code is available at: https://github.com/yelboudouri/EmoNeXt

\subsection{Results}
\label{results}
\begin{table}[htbp]
    \centering
    \caption{Performance comparison on FER2013 test set}
    \begin{tabular}{ll}
    \hline
    Model & Accuracy (\%) \\
    \hline
    GoogleNet \cite{giannopoulos2018deep} & 65.20 \\
    Deep Emotion \cite{minaee2021deep} & 70.02 \\
    Inception \cite{khaireddin2021facial} & 71.60 \\
    \textbf{ConvNeXt-Tiny} & \textbf{71.99} \\
    Ad-Corre \cite{fard2022ad} & 72.03 \\
    \textbf{ConvNeXt-Small} & \textbf{72.34} \\
    SE-Net50 \cite{khanzada2020facial} & 72.50 \\
    ResNet50 \cite{khanzada2020facial} & 73.20 \\
    \textbf{ConvNeXt-Base} & \textbf{73.22} \\
    VGG \cite{khaireddin2021facial} & 73.28 \\
    \textbf{EmoNeXt-Tiny} & \textbf{73.34} \\
    \textbf{ConvNeXt-Large}  & \textbf{73.46} \\
    Residual Masking Network \cite{pham2021facial} & 74.14 \\
     \textbf{ConvNeXt-XLarge}  & \textbf{74.15} \\
    LHC-NetC \cite{pecoraro2022local} & 74.28 \\
    \textbf{EmoNeXt-Small} & \textbf{74.33} \\
    LHC-Net \cite{pecoraro2022local} & 74.42 \\
    \textbf{EmoNeXt-Base} & \textbf{74.91} \\
    CNNs + BOVW \cite{georgescu2019local} & 75.42 \\
    \textbf{EmoNeXt-Large} & \textbf{75.57} \\
    Segmentation VGG-19 \cite{vignesh2023novel} & 75.97 \\
    \textbf{EmoNeXt-XLarge} & \textbf{76.12} \\
    \hline
    \end{tabular}
    \label{tab:accuracy_table}
\end{table}

The results presented in Table \ref{tab:accuracy_table} demonstrate the superior performance of our proposed model, EmoNeXt, compared to existing state-of-the-art single network architectures trained on the FER2013 dataset. Among the listed models, EmoNeXt-XLarge stands out with an accuracy of 76.12\%. This achievement can be attributed to the unique design and architecture of EmoNeXt, which effectively captures and emphasizes relevant facial features for accurate emotion classification.

When considering accuracy, EmoNeXt-Tiny achieves an accuracy of 73.34\%, surpassing well-known models like ResNet50 (73.20\%) and VGG (73.28\%), as well as the three first versions of ConvNeXt: tiny, small, and base.

With notable progress, EmoNeXt-Small exhibits enhanced performance compared to EmoNeXt-Tiny, attaining an accuracy of 74.33\%. This achievement surpasses advanced architectures like the Residual Masking Network (74.14\%) and LHC-NetC (74.28\%), as well as the last two sizes of the original ConvNext (Large and XLarge). These results underscore EmoNeXt-Small's exceptional ability to effectively capture and classify facial emotions.

EmoNeXt-Base maintains its performance by achieving an accuracy of 74.91\%, surpassing models like LHC-NetC (74.28\%). On the other hand, EmoNeXt-Large achieves an accuracy of 75.57\%, outperforming the combination of CNN and BOVW models (75.42\%).

Notably, EmoNeXt-XLarge achieves an accuracy of 76.12\%, surpassing the current best state-of-the-art accuracy attained by the Segmentation VGG-19 model (75.97\%). This remarkable performance firmly establishes EmoNeXt-XLarge as a highly effective model for image classification tasks, specifically for facial emotion recognition (FER).

\section{Conclusion}
In this paper, we presented EmoNeXt, a novel deep learning framework for facial expression recognition based on an adapted ConvNeXt architecture network. The EmoNeXt model integrates a Spatial Transformer Network (STN), Squeeze-and-Excitation blocks, and self-attention regularization to capture rich facial features and improve emotion classification accuracy. Experimental results on the FER2013 dataset demonstrate the superiority of EmoNeXt over existing state-of-the-art models. The integration of STN, SE blocks, and self-attention provides robust and accurate facial emotion detection, making EmoNeXt a promising approach for various applications requiring emotion recognition.

A detailed study is underway and will be published in a forthcoming journal paper. In this upcoming publication, our objective is to conduct a thorough evaluation of our model and compare its performance with other models, using diverse FER databases. 

Furthermore, we have future plans to extend the application of our model to emotion recognition in elderly Alzheimer's patients, highlighting its potential for real-world impact and further research avenues.

\section*{Acknowledgment}

This work was funded by the Dijon Metropole under contract DEVECO DM2023-029-20230301. The authors would like to thank Dr. Imad SFEIR from VYV3 Bourgogne for the helpful discussions.

\bibliographystyle{./IEEEtran}
\bibliography{IEEEabrv,IEEEexample}

\end{document}